\documentclass[10pt,twocolumn,letterpaper]{article}

\usepackage{cvpr}
\usepackage{times}
\usepackage{epsfig}
\usepackage{graphicx}
\usepackage{amsmath}
\usepackage{amssymb}

\usepackage{multicol}
\usepackage{multirow}
\usepackage{makecell}
\usepackage{booktabs}
\usepackage{color}
\usepackage{float}

\usepackage{amsthm,amsmath,amssymb}

\usepackage{mathrsfs}

\usepackage{subfigure}
\newcommand{\tabincell}[2]{\begin{tabular}{@{}#1@{}}#2\end{tabular}}  

\usepackage[breaklinks=true,bookmarks=false]{hyperref}

\cvprfinalcopy 

\def\cvprPaperID{****}

\begin{document}

\title{Semantic-Aware Pretraining for Dense Video Captioning}

\author{Teng Wang$^{1 2}$, Zhu Liu$^1$, Feng Zheng$^1$, Zhichao Lu$^1$, Ran Cheng$^1$, Ping Luo$^2$ \\
$^1$ Southern University of Science and Technology, $^2$ The University of Hong Kong\\
{\tt\small \{wangt2020,liuz2019\}@mail.sustech.edu.cn, \{zhengf,luzc,chengr\}@sustech.edu.cn, pluo@cs.hku.hk}}

\maketitle

\begin{abstract}
    This report describes the details of our approach for the event dense-captioning task in ActivityNet Challenge 2021\footnote{\url{http://activity-net.org/challenges/2021/tasks/anet_captioning.html}}. We present a semantic-aware pretraining method for dense video captioning, which empowers the learned features to recognize high-level semantic concepts. Diverse video features of different modalities are fed into an event captioning module to generate accurate and meaningful sentences. Our final ensemble model achieves a 10.00 METEOR score on the test set.
\end{abstract}

\section{Approach}
The goal of dense video captioning is to detect and describe all the events in an untrimmed video, which relies on rich spatial-temporal video features. We extract the clip-level features of the raw video by various encoders with different modalities and different pretraining tasks. The final video representation is obtained by the concatenation of selected features along the channel axis. Afterward, we adopt an off-the-shelf dense video captioning model~\cite{wang2020dense} to generate the locations and captions for multiple event proposals. Finally, we ensemble the generated captions from different models for a further performance boost.

\subsection{Semantic-Aware Pretraining}
Mainstream dense video captioning methods adopt video encoders pre-trained on action classification datasets, where action-oriented supervisions guide the model to the focus on the motion and appearance features related to a limited number of action classes. 
However, these methods fail to explicitly model the fine-grained semantic components, like objects, numbers, and colors, which are essential for caption generation. In this report, we propose a pre-training task named semantic concept classification~(SCC), serving as an auxiliary objective for a semantic-aware video feature extractor.

\begin{figure}
    \centering
    \includegraphics[width=0.5\textwidth]{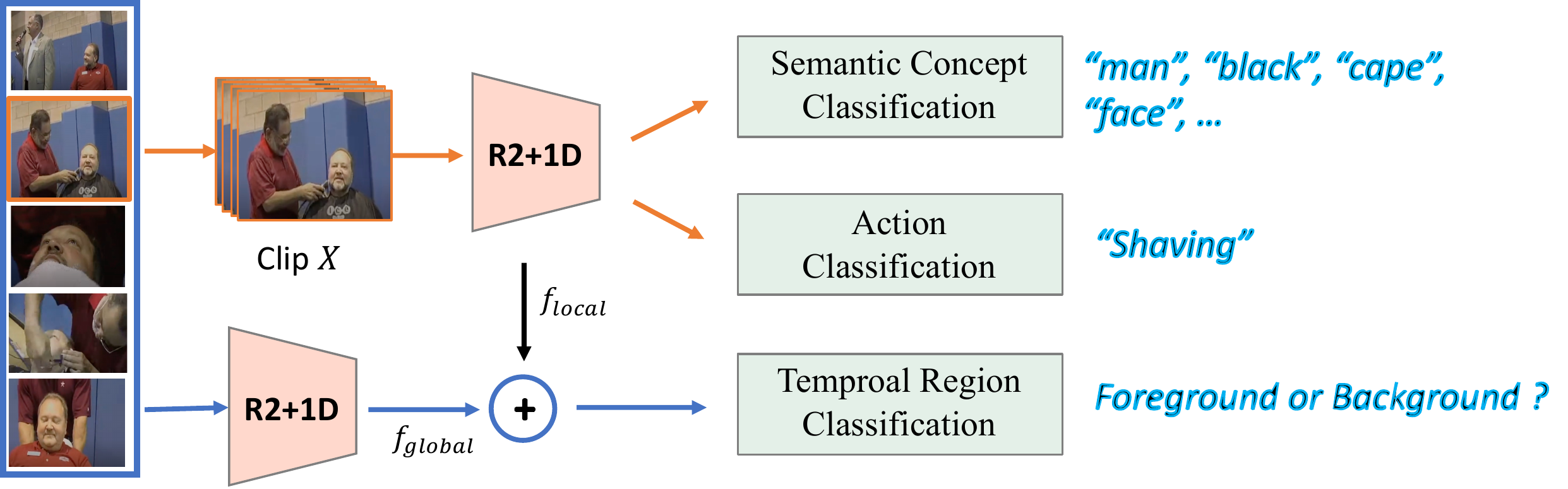}
    \caption{The proposed pretraining strategy. We use R(2+1)D to produce the local clip features of a clip and the global features of the entire video. Then the local and global features are leveraged to perform three pretraining tasks, i.e., action classification, semantic concept classification, and temporal region classification. By doing so, the learned representation is semantic-aware and temporal-sensitive. }
    \label{fig:sap}
\end{figure}

We build the video encoder based on TSP~\cite{alwassel2020tsp}, a supervised pretraining paradigm for temporally-sensitive representation learning for untrimmed videos. Given a video clip, the video encoder aims to predict 1) the foreground class of the clip (action classification), 2) whether the clip is in the foreground or the background of the video (temporal region classification), and 3) the semantic concepts in the clip. The overall pretraining objective is: 
\begin{equation}
L = L_{\rm action} + \alpha L_{\rm temporal} + \beta L_{\rm semantic},    
\label{eq1}
\end{equation}
where $L_{\rm action}$ and $L_{\rm temporal}$ represent the cross-entropy losses for action classification and temporal region classification, and $L_{\rm semantic}$ is the multi-label classification loss for SCC. We use R(2+1)D-34~\cite{tran2018closer} as the backbone of the video encoder. The predicted probability of semantic concepts is obtained by an FC+Sigmoid layer onto the local clip features encoded by R(2+1)D-34. 
The overall architecture is shown in Figure \ref{fig:sap}.

To construct semantic labels, we select the top $N$ frequent nouns, verbs, adjectives, and adverbs from the ground-truth sentences in ActivityNet Captions as the vocabulary of semantic concepts. Since most clips contain few positive semantic labels, the negative ones will dominate the loss and hurt the representation ability of the model. To tackle the positive-negative imbalance problem, we employ the asymmetric loss~\cite{ben2020asymmetric} as $L_{\rm semantic}$. 

\begin{table*}[th]
\begin{center}
\setlength{\tabcolsep}{1.2 mm}{
\small
\begin{tabular}{l c c c ccccc}
    \toprule
    Video Features &  Modality & Pretraining task & Pretraining Dataset  & METEOR & BLEU4 & CIDEr  \\
    
    \midrule
    VGGish~\cite{hershey2017cnn} & Audio & Audio Cls. & YouTube-8M &  9.23 & 1.72 & 35.49 \\
    ResNet152~\cite{he2016deep} & RGB & Image Cls. & ImageNet & 10.87 & 2.55 & 45.79 \\
    I3D~\cite{carreira2017quo} & RGB + Flow & Action Cls. & Kinetics & 11.43 & 2.79 & 49.77  \\
    TSN~\cite{xiong2016cuhk} & RGB + Flow & Action Cls. & Kinetics + Anet1.3 & 11.49 & 2.85 & 49.34 & \\
    SlowFast~\cite{feichtenhofer2019slowfast} & RGB & Action Cls. & Kinetics & 11.59 & 2.86  & 49.72\\
    TSP~\cite{alwassel2020tsp} & RGB &  Action Cls. + Temp. Cls. & IG65m +Kinetics + Anet1.3 & 11.56 & 2.93 & 50.68 \\
    \textbf{SA-TSP(ours)} & RGB  & Action Cls. + Temp. Cls. + Semantic Cls. & IG65m +Kinetics + Anet1.3  & \textbf{11.70} & \textbf{3.10} & \textbf{52.16}  \\
    \bottomrule
\end{tabular}}
\end{center}
\caption{Comparison of different pretraining strategies for video feature encoder. The METEOR/BLEU4/CIDEr scores with ground-truth proposals are calculated on the validation set by the official evaluation toolkit.}
\label{table:FeatureTypes}
\end{table*}

\begin{table}[th]
\begin{center}
\setlength{\tabcolsep}{1.2 mm}{
\small
\begin{tabular}{l c c }
    \toprule
    {Model} & \tabincell{c}{ METEOR \\ \scriptsize with GT proposals}  & \tabincell{c}{ METEOR \\ \scriptsize with learned proposals }  \\
    \midrule
    Feature combination & 12.28 & 8.16 \\
    + Enlarged training set & 12.33 & 8.03 \\
    + SCST & 15.89 & 11.42 \\
    + Model ensemble & \textbf{16.19}  & \textbf{11.50} \\
    \bottomrule
\end{tabular}}
\end{center}
\caption{Performance on the validation set.}
\label{table:trick2}
\end{table}

\subsection{Feature Combination}
To increase the diversity of video features, we extract different types of video features across various input modalities (including RGB, optical flow, and audio) and pretraining tasks (including image classification, audio classification, action classification, temporal region classification, and semantic concept classification).  Then, we find the best feature combination from the feature pool according to their captioning performance. The best feature combination is ``SA-TSP + I3D + VGGish".

\subsection{Model Ensemble}

We train $M$ event captioning models by varying the video feature combination strategy and the random seed. For each proposal, a caption set with a size of $M$ is generated and the best caption is selected by two criteria: the number of unique semantic concepts and the max inverse document frequency (IDF) of all unigrams.

\subsection{Implementation Details}
In the semantic-aware pretraining, we set the trade-off factors $\alpha$, $\beta$ in Eqn.~(\ref{eq1}) to be 1 and 0.1, respectively. The size of semantic concept vocabulary $N$ is 1000. We first train the R(2+1)D backbone on the IG65m+Kinetics dataset for action classification, then finetune the model based on the proposed pretraining objective in Eqn.~(\ref{eq1}) on the ActivityNet1.3 dataset.

For temporal action proposals, we directly use the predicted results provided by~\cite{wang2020dense}, achieving 40.09\% and 66.63\% in terms of precision and recall on the ActivityNet validation set. For event captioning, we use a vocabulary with a size of 8340. The size of all hidden layers is set to 512. We adopt the Adam optimizer with an initial learning rate of 1e-4. The batch size is set to 1 and the max training epoch is 40. The number of ensemble models $M$ is~5.

\section{Evaluation Results}

Table~\ref{table:FeatureTypes} shows the captioning performance with a single type of feature. We observe that the similarity between the pretraining task and downstream task matters. The unsatisfying performance of VGGish and ResNet152 is probably caused by the large discrepancy between the audio/image classification and the dense video captioning. The proposed SA-TSP achieves the best results among the eight models, which verifies the effectiveness of the semantic concept classification loss.

Table~\ref{table:trick2} shows several techniques to boost the dense video captioning system. Feature combination yields a clear performance gain (11.70$\rightarrow$12.28) compared with the single-model performance. We further extend the official train set by appending around 80\% of the videos from the validation set. Then we finetune the event captioning module with the self-critical sequence training (SCST)~\cite{rennie2017self} on the enlarged train set, which gives a considerable performance gain both with ground-truth proposals and with learned proposals. The final submission is obtained by ensembling five different event captioning models. On the test set, our final submission achieves a 10.00 METEOR score.

{\small
\bibliographystyle{ieee}
\bibliography{egpaper_final}
}

\end{document}